# A Decision Making Approach for Chemotherapy Planning based on Evolutionary Processing

Mina Jafari, Behnam Ghavami, Vahid Sattari Naeini
Department of Computer Engineering, Shahid Bahonar University of Kerman
Kerman, Iran
m.jafari@eng.uk.ac.ir, ghavami@uk.ac.ir, vsattari@uk.ac.ir

*Abstract*— The problem of chemotherapy treatment optimization can be defined in order to minimize the size of the tumor without endangering the patient's health; therefore, chemotherapy requires to achieve a number of objectives, simultaneously. For this reason, the optimization problem turns to a multi-objective problem. In this paper, a multi-objective meta-heuristic method is provided for cancer chemotherapy with the aim of balancing between two objectives: the amount of toxicity and the number of cancerous cells. The proposed method uses mathematical models in order to measure the drug concentration, tumor growth and the amount of toxicity. This method utilizes a Multi-Objective Particle Swarm Optimization (MOPSO) algorithm to optimize cancer chemotherapy plan using cell-cycle specific drugs. The proposed method can be a good model for personalized medicine as it returns a set of solutions as output that have balanced between different objectives and provided the possibility to choose the most appropriate therapeutic plan based on some information about the status of the patient. Experimental results confirm that the proposed method is able to explore the search space efficiently in order to find out the suitable treatment plan with minimal side effects. This main objective is provided using a desirable designing of chemotherapy drugs and controlling the injection dose. Moreover, results show that the proposed method achieve to a better therapeutic performance compared to a more recent similar method [1].

I. INTRODUCTION

Cancer occurs when a certain set of genes that is responsible for normal cell division leads to failure; they activated or altered at the wrong time; hence, cells grow out of control and form a tumor. For instance, the gene that controls cell division is damaged by mutation automatically. As a result of this damage, changes occur in the ordinary process of division [2]. Surgery, chemotherapy, radiation therapy, hormone therapy and immunotherapy are cancer treatment options. Since cancer cells invade to the surrounding tissues and migrate to other parts of the body, chemotherapy is usually selected as a systemic treatment by physicians; however, the effectiveness of the chemotherapy is limited due to the drug toxicity and drug interactions with normal cells and other tissues. Thus, there is a need of designing an appropriate drug regimen to balance between the effectiveness of the treatment (destroying the cancerous cells) and toxic side effects. Currently, evaluating the efficacy and the amount of toxicity are done based on empirical evidence obtained from clinical trials; however, clinical trials are not desirable protocols due to the limited human and financial resources. For this purpose, mathematical models as low-cost ways have been welcomed to assess the impact of drugs effectiveness [1, 3].

Each person has a different reaction to a drug; varied reactions come from slight changes in people's genes. In some cases, these changes show that people need different doses of a drug; in other cases, suggesting that drugs, which are effective for a person, are fatal to another one. Since each person has unique genetic information, it can be concluded that genetic diversity causes each person responds differently to the same drug; moreover, it brings about the fact that different side effects appear. In order to reduce the side effects of drugs and increase effectiveness of cancer treatment and other diseases associated with genes, there is a need of treatment based on genetic information that is known as personalized medicine [4, 5]. Personalized medicine is an emerging medical process that utilizes people genetic profile to guide decisions about prevention, diagnosis and treatment of genetic diseases such as cancer. The awareness of genetic information of a patient helps physicians about choosing appropriate drug or treatment plan using suitable drug dose or regime [6-9].

A therapeutic plan using chemotherapy will be considered with the aim of reducing the number of cancerous cells (tumor size), periodically. Considering that drugs inject into the body through the blood, chemotherapy acts such as a double-edged sword; on one hand, it destroys cancer cells. On the other hand, it destroys the normal and healthy cells and enters toxicity into the body that causes many side effects. Therefore, the goals of chemotherapy are minimizing the number of cancer cells and reducing the toxicity entering to the body. These two objectives are contradictory; thus, it is necessary to design an effective and optimal chemotherapy plan [8, 9]. In chemotherapy, according to the patient's status and the growth stage of the tumor, the goals of treatment are changed and are given below [10]:
- Cure: In this phase, the aim is to destroy cancer cells.
- Control: If cure is not possible, the main focus is controlling the cancer. In fact, in this phase the aim is to prevent the spread of cancerous cells in the body.
- Palliation: If the cancer is in an advanced stage and the two aforementioned goals cannot be applied; therefore, the aim is to enhance the quality of patient's life. (An idea in this step is reducing the amount of drugs in order to reduce the toxicity).

As mentioned above, to find an optimal schedule of chemotherapy, various mathematical models have been proposed in which tumor growth is calculated by taking the effect of their medications into account. In fact, mathematical models are provided to simulate the tumor response after the specified period of time using the prescription drugs. Models that consider the cell cycle act regarding cell behaviour. Cell cycle is a chain of steps that has an effect on both normal and cancer cells, from their birth until their death. In general, cell cycle consists of five stages: resting phase or zero Gap (G0), first Gap (G1), Synthesis (S), the second Gap (G2), and mitosis (M). Cell cycle is shown in Figure 1. Usually, anti-cancer drugs stop dividing task using RNA or DNA damaging [11]. Anti-cancer drugs that are able to kill all cancer cells called cell cycle non-specific drugs; while drugs that only have the ability to kill cancerous cells that are dividing called cell-cycle specific drugs [1].

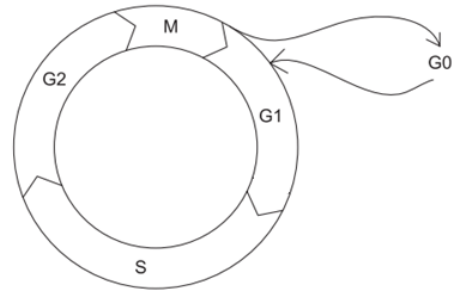

**Figure 1: Cell cycle [12]**

Researchers pay special attention to the first day of treatment [13]. As a matter of fact, it is imperative to prescribe the strongest treatment plans that destroy the greatest number of cancer cells because on the first day of treatment the body has a high metabolism capacity; the body still has not shown resistance to drugs. Moreover, considering immune system [14] and body resistance [15] to drugs in the chemotherapy scheduling are the issues that have been studied.

Chemotherapy drugs are usually delivered to the body based on a discrete dose. For instance, if there are $n$ doses, drugs are injected to the body at times $t_1, t_2, ..., t_n$. Where multiple drugs are used instead of only one drug, usually a combination of these drugs are given to the patient's body at each time. If there are $d$ different drugs, the concentration of the drugs at each time are determined as follows [10, 16]:

$$C_{ij}, i \in \overline{1,n}, j \in \overline{1,d} \quad (1)$$

where $C_{ij}$ determines the drug concentration in $i$th time using $j$th drug.

In this paper, a method is proposed to reduce the side effects of chemotherapy drugs and it chooses the most suitable chemotherapy treatment plan using a meta-heuristic Multi Objective Particle Swarm Optimization (MOPSO) algorithm. The amount of drug that should be injected to a patient's body is coded in each particle. Next, the drug concentration is calculated and in the next step, the entered toxicity and tumor size are determined. The proposed method returns a set of solutions as output that phisicans can choose a solution based on three objectives, i.e. cure, control and palliation.

The rest of this paper is organized as follows. In section II, the related works are presented. In section III, the proposed method to optimize cancer chemotherapy plan based on MOPSO is described. The experimental results of the proposed method are expressed in section IV and finally, section V concludes the paper.

II. Related Works

Recently many methods have been proposed to optimize chemotherapy plan [17, 18]. The effectiveness of each method is measured according to tumor size (the number of cancerous cells) and the toxicity entered to the body. In this section, some of the proposed methods are expressed for optimal design of chemotherapy plan.

Petrovski and McCall [19] have utilized Strength Pareto Evolutionary Algorithm (SPEA) in order to optimize the problem of chemotherapy. In their method, the most efficient treatment plan is obtained by changing the drug concentration; therefore in this way, the search space is a set of vectors that show the drug concentration. To predict the tumor response to chemotherapy, a mathematical model that is called the Gompertz growth model is used. The first objective function is introduced to reduce the number of cancerous cells and the second one is selected to increase Patient Survival Time (PST).

Petrovski et. al [20] have compared two Genetic Algorithm (GA) and Particle Swarm Optimization (PSO) algorithms in order to choose the optimal treatment plan. In their work, each chromosome is a set of concentrations of different drugs with different doses and the objective function is considered according to the Gompertz model to reduce the number of cancerous cells. The results showed that the PSO algorithm is faster than GA to find the optimal response in the space of treatment solutions.

Liang et. al [13] have provided several optimal chemotherapy plans using a multi-modal algorithm which makes it possible to choose a suitable plan for each patient, in order to apply goals of personalized medicine. In their investigation, the objective function is expressed using the Gompertz mathematical model.

Tse et. al [15] have introduced an effective model for chemotherapy using multiple drugs. Three drugs have been used and a new Memetic Algorithms (MA) is provided using Iterative Dynamic Programming (IDP) and Adaptive Elitist Genetic Algorithm (AEGA) to determine the best treatment plan. In their paper, the tumor is composed of many different parts; each part is resistant to certain drugs. Mathematical models have been considered to assess tumor size using three drugs. The advantage of their technique is considering the body's resistance to drugs in chemotherapy process.

Petrovski et. al [10] have introduced an optimal chemotherapy plan using a MOPSO algorithm. The first objective function has been considered to minimize the number of cancerous cells. The second objective function is intended to control cancer. Objective functions are considered based on Gompertz model. Each particle is in the form of numbers which are indicative of the concentration of drugs.

A multi-objective optimization based on smart PSO with decomposition is proposed by Moubayed et. al to optimize chemotherapy plans [16]. In fact, this approach is based on a PSO algorithm to parse a multi-objective optimization problem on several problems, in order to reduce the complexity. The first objective function is modelled to reduce the size of the tumor and the second one is introduced to minimize the toxicity.

Alam et. al [21] have introduced an optimal scheduling plan in order to control the dose of chemotherapy drugs that must be injected into the patient's body during a period. In their method, a multi-objective GA is used to select the most appropriate drug concentration which balances between reducing the toxicity and minimizing the number of cancerous cells. The first objective function is considered to minimize the number of cancerous cells and the second one is introduced to reduce the amount of toxicity during a cycle of treatment. The proposed algorithm returns a wide range of solutions as output which physicians can choose the most appropriate solution for each person based on disease and cancer growth stage.

Ochoa and Villasana [14] have studied on the suitability population-based algorithms in order to provide the best chemotherapy plans. In their investigation, three methods are studied, the Covariance Matrix Adaptation (CMA) evolution strategy, Differential Evolution (DE) and PSO. All three algorithms are able to achieve success in solving the problem posed while the DE has better performance than their counterparts overall. This method is provided to take cancerous cells in the different stages of the cell cycle and immune response into account. The power of their method is using the mathematical models that capable of tracking and measuring the cells in different phases of cell cycle. In addition, the immune system is also considered which is a crucial issue in chemotherapy.

An optimal control model to optimize the chemotherapy planning problem during a cycle using an anticancer drug (VP-16) have provided by Zhu et. al [1]. In their method, three mathematical models are used to mimic the physiological response of the body that is affected by chemotherapy: pharmacokinetics (PK) model which represents the distribution of anti-cancer drug, A two compartments dynamic model that investigate the effect of cell cycle-specific anticancer drugs on tumor growth and a semi-mechanical model is intended to study the bone marrow suppression (toxicity entered to the body). The toxicity entered into the body is calculated by the numbers of white blood cells. If this number is less than a threshold, it indicates there is a large amount of drug's toxicity. This method is very close to reality in terms of clinical issues. For instance, considering the number of white blood cells as a measure to assess the toxicity entered to the body is one of the benefits of their approach.

To sum up, one of the advantages of the Gompertz model is that this model is known as one of the most appropriate models for describing tumor growth and it is the most used model in the literature. However, there are restrictions on the use of the Gompertz model to describe the population of cancerous cells affected by chemotherapy especially when anticancer drugs are cell-cycle specific. Since this class of drugs have an effect on tumor cells only in certain phases; thus, this is not an appropriate model to simulate responses of these types of drugs. Since it is not efficient to investigate the influence of a drug at different phases of the cell cycle [1]. In this paper, we have provided a more practical approach in order to reduce the gap between theoretical research and the medical profession. The objective functions in the proposed method are obtained using the mathematical models presented in [1] which are very close to clinical issues.

### III. BACKGROUND: PARTICLE SWARM OPTIMIZATION ALGORITHM

Particle Swarm Optimization (PSO) is one of the intelligent optimization algorithms in the field of swarm intelligence, which is introduced by James Kennedy and Russell C. Eberhart [22]. This algorithm inspired by the social behavior of animals such as fish and birds, which live together in large and small groups. In this algorithm, members of the population (solutions) communicate with each other directly and solve the problems through the exchange of information with each other and remembering the past positions [23]. In this algorithm, all solutions are available with a value of merit, which is obtained using a fitness function. The purpose of this algorithm is finding where the best value of fitness function is in the search space. The amount of fitness function in the direction and velocity of the solution has a direct impact to find the optimal solution. The pseudo-code of this algorithm is shown in Figure 2.

---
**PSO Algorithm**
- Generate initial population randomly.
- Initialize the position and velocity of each particle
- Calculate fitness values corresponding to each particle in population
- Calculate PBest and GBest for each particle
- Repeat
  - Update velocity and position of each particle
  - Calculate fitness values corresponding to each particle in population
  - Update PBest and GBest for each particle
- Until termination condition is satisfied
---

**Figure 2: The pseudo-code of PSO algorithm**

This algorithm starts working by using a population of initial solutions and moving these solutions over successive iterations; it tries to find the optimal solution. In each iteration, *GBest* and *PBest* are updated where *PBest* equal with position the best fitness value of each particle during its movement so far and *GBest* is the position of the best particle in entire population. After finding the above values, the velocity of particles are calculated by equation 2 and the position of each particle is obtained as equation 3 [24]:

$$V_{i,t+1} = W * V_{i,t-1} + C_1 * r_1(PBest_i - P_{i,t}) + C_2 * r_2(GBest_i - P_{i,t}) \qquad (2)$$
$$P_{t+1} = P_t + V_t \qquad (3)$$

where $r_1$ and $r_2$ are random values in the range of [0,1], $C_1$, $C_2$ are coefficients and $W$ is the weight of inertia.

Many real-world problems are associated with more than one objective. These kinds of problems have been identified as multi-objective problems[25]. The solution of multi-objective problems is establishing the best possible negotiation among all the objectives; thus, a multi-objective problem can include uncountable set of solutions that compromise among the various objectives. In a multi-objective optimization problem, there are *n* decision variables that form decision vector $X = x_1, x_2, ..., x_n$, there are *m* constraints $g_1(x), g_2(x), ..., g_m(x)$ and *k* objective functions $f_1(x), f_2(x), ..., f_k(x)$. PSO mainly due to the high speed of convergence is very suitable for multi-objective optimization problems [26, 27].

To this day, a set of ideas are provided in the literature to change the typical PSO to multi-objective version. In this paper, the model presented in [26] is used. In this method, a hypercube is generated based on the search space explored so far and particles are located according to their fitness functions. An external repository is used entitled *Rep*. From the non-dominated set a leader is selected (a particle) to guide the particles toward the optimal Pareto front. It is obvious that the choice of a leader is a key concept for the design MOPSO. Each solution in non-dominated set can be considered as a new leader; from all of them, only one particle introduce as new leader. Another complex task in MOPSO is updating the external archive. If the new solution is not dominated by all of the archive members, the solution will be added in external archives. In addition, if the new solution dominates some members of the external archive, dominated solutions will be deleted. To avoid from the complexity of updating archive; its size is considered limited. Since *Rep* size is limited, in emergency cases the algorithm gives a priority to the particles that are located in less space populated. *Rep* memory contents and the position particles in the hypercubes are updated, simultaneously.

### IV. EFFECTIVE CHEMOTHERAPY PLANNING USING MOPSO ALGORITHM

In this section an efficient method to optimize the chemotherapy plan will be presented based on MOPSO. First, we present the mathematical model and next, the proposed algorithm will be described in detail.

## A. Mathematical Model

Using the PK model, the drug distribution by the circulatory system will be determined and eventually the drug concentration will be specified in plasma [28]. Because the drug that is injected to the body can transfer to other parts in addition to the plasma; this model is intended as a dual compartment. The model is shown in Figure 3.

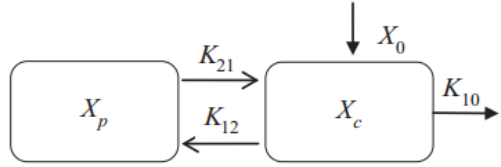

Figure 3: The two compartment PK model [1]

The amount of drug in plasma $X_c$, in peripherals $X_p$ and the drug concentration $C$ are calculated based on mathematical equations as follows [1]:

$$\dot{X}_c = K_{21} * X_p - (K_{12} + K_{10}) * X_c + X_0 \tag{4}$$

$$\dot{X}_p = K_{12} * X_c - K_{21} * X_p \tag{5}$$

$$C = X_c / V_c \tag{6}$$

$K_{12}$ and $K_{21}$ are equal to the rate of drug transfer between the two copartment $X_c$ and $X_p$. $K_{10}$ is equal to the rate of drug clearance, $X_0$ drug dose, $V_c$ central compartment volume (plasma) and $C$ represents the concentration of drug in the plasma.

After calculating the concentration of the drug in the body, the tumor growth can be calculated. In this paper, the model of tumor growth is intended in the form of a two compartment model as shown in Figure 4. Resting cells are the ones that are not affected directly from the anti-cancer drug and the cycling cells are the ones that are directly affected by the drug. In fact, cells in the cycling cells compartment are containing cells in phases $G_1$, $S$, $G_2$ and $M$ (four compartments are shown in a compartment) and the cells in $G_0$ are considered as resting cells compartment.

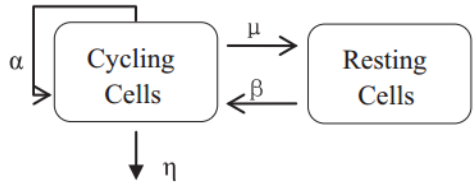

Figure 4: Tumor growth model [1]

The cycling cells $X_1$ and resting cells $X_2$ are calculated using the following equations [1]:

$$\dot{X}_1 = (\alpha - \mu - \eta) * X_1 + \beta * X_2 - g(t) * X_1 \tag{7}$$

$$\dot{X}_2 = \mu * X_1 - \beta * X_2 \tag{8}$$

where $\alpha$ is cell growth rate in cycling cells compartment, $\mu$ and $\beta$ are mutation rate between the two compartment $X_1$ and $X_2$, $\eta$ is the natural decay rate of the cycling cells compartment and also natural decay rate is considered zero for resting cells compartment [29]. Moreover, $g(t)$ is the effect of the drug on cancerous cells and is calculated as [1]:

$$g(t) = K_1 * C \tag{9}$$

The Pharmacodynamic (PD) model is described to calculate the toxicity of the drug into the body according to a semi-mechanical model as shown in Figure 5.

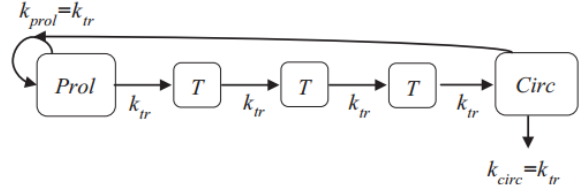
**Figure 5: Semi-mechanical model for the study of bone marrow suppression [1]**

Mathematical equations of the above model are calculated as follows [1]:

$$\dot{Prol} = K_{prol} * Prol * (1 - E_{drug}) * (Circ_0/Circ)^\gamma - K_{tr} * Prol \qquad (10)$$
$$\dot{T}_1 = K_{tr} * Prol - K_{tr} * T_1 \qquad (11)$$
$$\dot{T}_2 = K_{tr} * T_1 - K_{tr} * T_2 \qquad (12)$$
$$\dot{T}_3 = K_{tr} * T_2 - K_{tr} * T_3 \qquad (13)$$
$$E_{drug} = Slope * C \qquad (14)$$
$$\dot{Circ} = K_{tr} * T_3 - K_{tr} * Circ \qquad (15)$$

where *Prol* represents the proliferation of bone marrow cells and three transmission casing $T_1$, $T_2$ and $T_3$ mimic white blood cells and neutrophils chain maturity. This chain delays the toxic effects of chemotherapy from the bone marrow to circulating white blood cells which is related to *Circ* in this model. $K_{prol}$ is the rate of production of new cells in *Prol*. $K_{tr}$ is intended to predict the the time delay between the implementation and the effect observed. $Circ_0$ is the initial number of white blood cells [30].

### B. The Proposed Algorithm

The proposed method uses MOPSO algorithm and provides the most appropriate chemotherapy plan in terms of minimizing the number of cancerous cells and also minimizing the toxicity entered to the body using a cell cycle-specific anticancer drug (VP-16). The $X_0$ determines drug dose and these values are encoded in each particle; thus, each particle is represented as follows:

$$X_0^i, i \in \{1,2,\ldots,n\}$$

where *n* is the number of times that the drug is delivered by the body. Figure 6 shows how to code a particle.

| $X_0^1$ | $X_0^2$ | $X_0^3$ | …….. | $X_0^{n-1}$ | $X_0^n$ |

**Figure 6: Codification of a particle in MOPSO**

The initial value of number of tumor cells is considered $1*10^{12}$ [1, 29]. Moreover, according to [1, 31], the initial value of $X_1$ is considered $8*10^{11}$ and also the initial value of $X_2$ is equal to $2*10^{11}$. It is necessary to eliminate the tumor completely or at least its size should be reduced as small as possible at the end of the treatment period. It should be noted that even one remaining cell may provoke recurrence of the disease [32]. On the other hand, the complete eradication of the tumor is not possible using by just chemotherapy. The toxicity entered to the body during the whole period of treatment should be considered low as much as possible; therefore, at the end of the treatment it is expected that the number of white blood cells become more than a specific threshold which indicates that the result of drug toxicity are not too much. In this paper, the neutrophil counts should be greater than the threshold $Circ_{th}$; that $Circ_{th}$ equal to $1.5 * 10^9$ $L^{-1}$ and $1*10^9$ $L^{-1}$, for levels two and three neutropenia, respectively [1]. Due to the risk of toxic side effects of chemotherapy, the dose of a drug that is injected into the patient's body should be low as much as possible and effective.

In the proposed method, the first objective function is considered to reduce the number of cancer cells as follows that this amount should be minimized:

$$minimize \quad F_1(X_0) = X_1 + X_2 \qquad (16)$$

In order to consider the toxicity of the drug and minimize it, the second objective function is intended in the form of the following equation. It is done by measuring the number of white blood cells *Circ* that this amount should be maximized:

$$maximize \quad F_2(X_0) = Circ \qquad (17)$$

The pseudo-code of the proposed algorithm is shown in Figure 7. To calculate the velocity of each particle, the following equation is used [26]:

$$V_{i,t+1} = W * V_{i,t-1} + r_1(PBest_i - P_{i,t}) \quad (18)$$
$$+ r_2(Rep_h - P_{i,t})$$

where the index $h$ is selected in this way: the hypercube that contains more than one particle, the assessment value of the hypercube is equal to the division of a number greater than 1 to the total number of particles in the hypercube. In fact, the objective is to reduce the assessment value of the hypercube that contains a large number of particles. The roulette wheel selection method is used to select the hypercube and then randomly a particle is chosen from the candidates, which is utilized as the leader.

When a decision variable goes beyond its scope one of the following rules are fired:
- The decision variable achieves a value equal to the its respective range (the minimum or maximum range).
- The velocity multiplied in -1 that causes it to begin search in the opposite direction.

The proposed method is very close to reality in terms of clinical issues. Considering the number of white blood cells as a measure to assess the toxicity entered to the body is one of the advantages of this method. Another advantage of this method is that the growth model capable of simulating the effect of drugs on cells in specific phases of the cell cycle, and anti-cancer drugs such as VP-16 can be modeled highly effective since it is a drug which only in the specific phases has an effects on a person's body. Also in this method the drug distribution is intended like a two-compartment model which calculates the effect of the drug on plasma and pheripherals. Another key point in the proposed method is using a powerful optimization algorithm (i.e. MOPSO) and the appropriate objective functions.

**Proposed Algorithm**
- Generate initial population randomly.
-Initialize the velocity of each particle
-Calculate fitness values corresponding to each particle in population based on Eq. 16 and 17
-Store the positions of the particles that represent non-dominated vectors in the second repository
-Generate hypercube of the search space explored so far and locate the particles according to the values of their objective functions
-Initialize the memory of each particle
-Repeat
- Compute the velocity of each particle based on Eq. 18
- Compute the new positions of the particles
- Use a strategy to maintain the particles within the search space scope
- Calculate fitness values corresponding to each particle in population
- Update the second repository and the location each particle in the hypercube
- The position of particle updates if the current position of the particle is better than the position contained in its memory
-Until termination condition is satisfied
-Return best population

**Figure 7: The pseudo-code of the proposed algorithm**

## V. EXPERIMENTAL RESULTS AND DISCUSSION

Chemotherapy involves a cycle of treatment that is usually considered in the form of three or four weeks. The reason is that the destroying of the cells that remain at rest during the previous treatment plans. In order to provide a quantitative assessment of MOPSO performance, three concepts are considered:
- Particle distribution,
- Spread particles across the optimal Pareto front,
- The ability to achieve a balance between two objectives.

We also investigate these three issues as well as the results of the proposed method which are compared with the one presented in [1]. The proposed method is implemented in MATLAB. The whole analysis is done using a 2.50 GHz Intel Core i5 processor with 4GB RAM.

MOPSO begins with a population of particles that are randomly generated. In this paper, the initial population is generated by 1,000 particles while in each particle 21 drug dose is stored (a three-week treatment period). The drug dose is variable in the range [0, 5] mg / $m^2$. MOPSO process runs for 100 generations to satisfy both objectives, simultaneously. Table 1 shows the parameter values used in the mathematical models in section IV. These parameters have been initialized based on the values reported [1, 28-31].

Table 1: Parameter values used in mathematical models

| Parameter | Value | Parameter | Value |
|---|---|---|---|
| $K_{12}$ | 0.14 | $\eta$ | 0.477 |
| $K_{21}$ | 0.06 | $\beta$ | 0.05 |
| $K_{10}$ | 1.14 | $\gamma$ | 0.17 |
| $V_c$ | 6 | Slope | 0.126 |
| $K_{tr}$ | 0.7680 | $X_p$ | 0 |
| $Circ_0$ | $5*10^9$ | $X_2$ | $8*10^{11}$ |
| $X_c$ | 0 | $\alpha$ | 0.5 |
| $X_1$ | $2*10^{11}$ | $\mu$ | 0.218 |
| $K_1$ | 0.03 | | |

Figure 8 shows the distribution of non-dominated solutions at the end of the experiment. The horizontal axis represents the number of tumor cells and the vertical axis represents the neutrophil counts. As seen in the figure, non-dominated solutions are well distributed. According to figure 8, for patients with the better physical conditions, dosing schedules from area 1 can be used to kill more tumor cells on the other hand these schedules enter more toxicity into the patient's body. Moreover, the number of white blood cells in this area is much reduced (cure) and in some circumstances the number of white blood cells are also less than the considered threshold. Dosing schedules from area 3 are considered for palliative chemotherapy and improves the quality of life of patients because it enters very little toxicity to the body and destroys lower the number of tumor cells (palliation). While dosing schedules from area 2 represent the balance between efficacy and side effects of anti-cancer drugs (control). In fact, it can be concluded that treatment programs are scheduled according to the amount $X_0$ which is the amount of the drug prescribed for a patient; if the drug dose is low, the antitumor effect is low too and in contradictory excessive drug doses reduces white blood cells and tumor cells too.

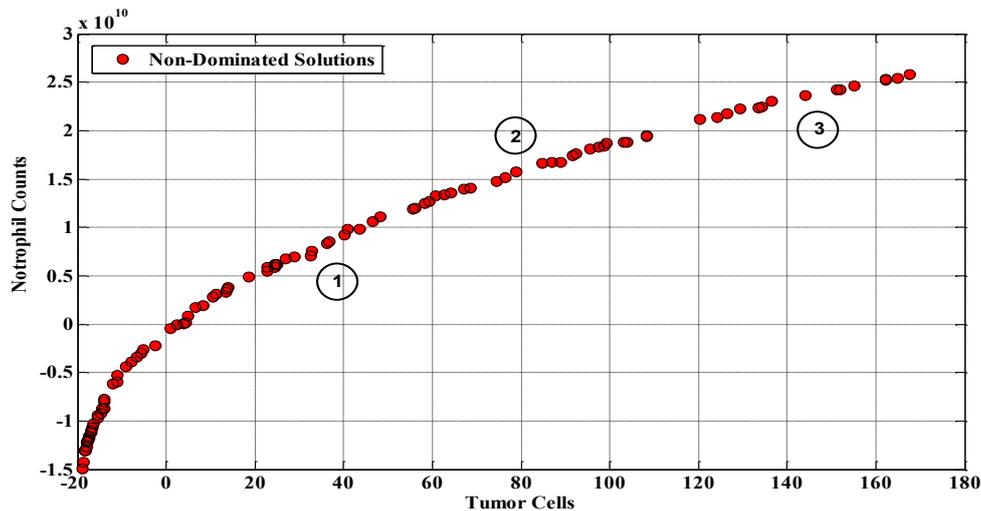

Figure 8: The distribution of non-dominated solutions in proposed method

To verify the effectiveness of the chemotherapy plans using the proposed method, some representative solutions are evaluated deeply. For this purpose, three solutions of the Pareto front are selected, one from each area. Sol-1 is chosen from area one, Sol-2 and Sol-3 are chosen from areas 2 and 3, respectively. Figure 9 shows the drug doses injected into the patient's body based on three selected solutions. Table 2 shows the average amount of injected drugs, concentration, and the average number of white blood cells to destroy tumor cells using three selected solutions. Figure 10 reveals changes the number of tumor cells during 21 days of treatment. In order to present the number of tumor cells better, this figure just reveals the number of tumor cells from 10*th* day to 21*th* day.

As shown in Table 2 and Figures 9 and 10, the Sol-1 is a solution which in the case of killing tumor cells are more succeful than the other two solutions at the end of the treatment period; therefore, it is expected that average drug dose entered be more than other ones and Sol-3 is a solution that the fewer drug dose enteres to the patient's body and against the number of tumor cells remain more.

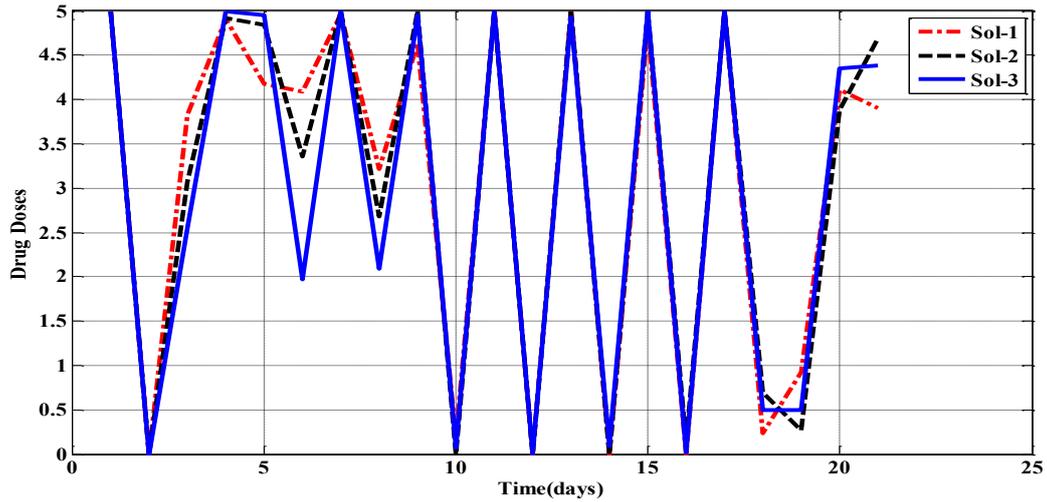

Figure 9: The drug dose entered into the patient's body during 21 days in three selected solutions

Table 2: the measurement performance Selected solutions at the end of 21 days.

| Selected Solutions | Drug Dose | Drug Concentration | Neutrophil Count | Reduction of cancerous cells | |
|---|---|---|---|---|---|
| | Avg | Avg | Avg | Cells Remaining | % Reduction |
| Sol-1 | 3.0480 | 4.3201 | 7.6436e+08 | 48 | ≈100 |
| Sol-2 | 3.0250 | 4.4131 | 1.0763e+09 | 96 | ≈100 |
| Sol-3 | 2.9216 | 4.5043 | 1.3995e+09 | 165 | ≈100 |

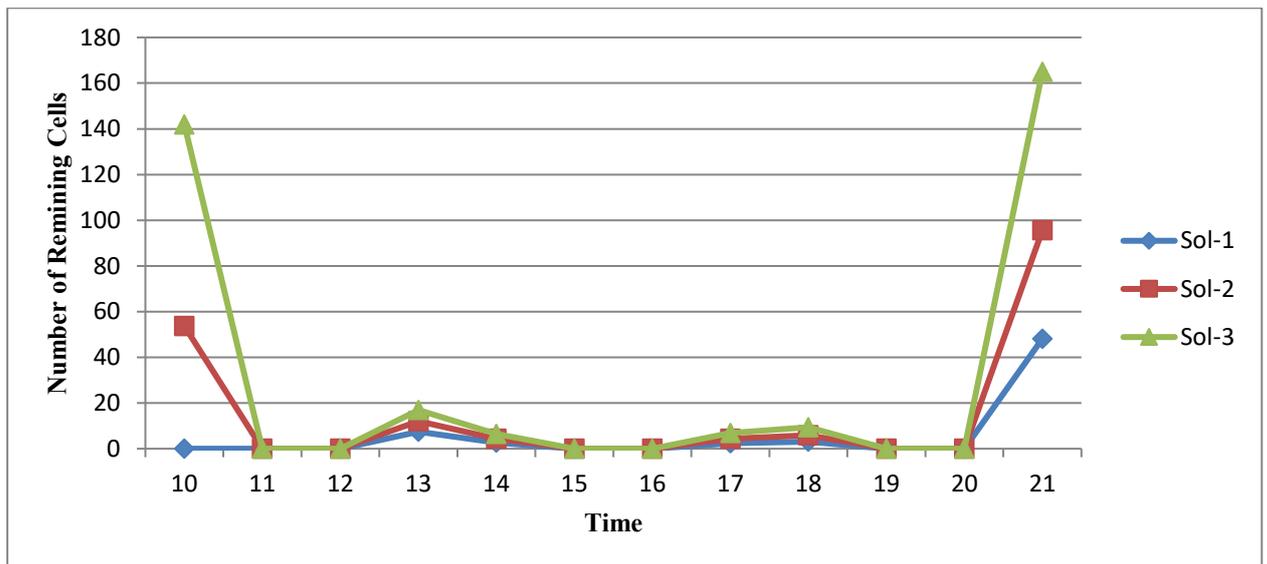

Figure 10: The changes in tumor size over a cycle of treatment using three selected plans

To study the efficacy of the proposed method compared to other methods, the proposed technique is compared with the presented method in [1]. As shown in Figure 8, the number of tumor cells in the proposed method varies in the range of [0,180] by changing the dose injected into the patient's body. While the number of tumor cells in [1] varies from $5.5*10^{11}$ to $10*10^{11}$ which confirm that the tumor size is nearly decreased 100% in the proposed method respect to [1]. The number of white blood cells in [1] varies to $5.5*10^{9}$ while this number is variable to $3*10^{10}$ in the proposed method that shows the toxicity entered into the body is less. Moreover, this method is compared to other methods that in all these methods which the number of tumor cells is calculated after 84 days of chemotherapy. This comparison is shown in Table 3.

Table 3 – the comparision among the proposed method at the end of 21 days and six other methods at the end of 84 days.

| Techniques | Number of Cells remain |
|---|---|
| Martin[33] | $4.878*10^4$ |
| Bojkov[34] | $3.31*10^4$ |
| Luus[35] | $2.57*10^4$ |
| Carrasco and Banga[36] | $1.534*10^4$ |
| Tan et al.[37] | $1.534*10^4$ |
| Liang et al.[38] | $1.760*10^3$ |
| The proposed method | 0-180 |

VI. CONCLUSION

In this paper, a multi-objective model for optimizing chemotherapy planning is proposed in order to reduce the cancerous cells with minimum side effects, after a specific treatment course. In the proposed method two objectives and some constraints are defined before starting the optimization process and the output is a wide range of perfect solutions. These solutions minimize the cancerous cells as well as side effects. The proposed method contains a variety of choices based on the patient's physiological condition. The experimental results reveal that the most of these solutions can extensively reduce cancerous cells compared to other methods. Moreover, the average toxicity and drug dose are at a low level.

REFRENCES